\documentclass[conference]{IEEEtran}

\usepackage[english]{babel}
\usepackage[utf8x]{inputenc}
\usepackage[T1]{fontenc}
\newcommand{\tabitem}{~~\llap{\textbullet}~~}

\usepackage{todonotes}

\title{Golden Years, Golden Shores: A Study of Elders in Online Travel Communities}
\author{\IEEEauthorblockN{Bart{\l}omiej Balcerzak, Rados{\l}aw Nielek}
\IEEEauthorblockA{Polish-Japanese Institute of Information Technology,\\ 
              ul. Koszykowa 86, 02-008 Warsaw, Poland\\}}
\begin{document}
\maketitle

\begin{abstract}
In this paper we present our exploratory findings related to extracting knowledge and experiences from a community of senior tourists. By using tools of qualitative analysis as well as review of literature, we managed to verify a set of hypotheses related to the content created by senior tourists when participating in on-line communities. We also produced a codebook, representing various themes one may encounter in such communities. This codebook, derived from our own qualitative research, as well a literature review will serve as a basis for further development of automated tools of knowledge extraction. We also managed to find that older adults more often than other poster in tourists forums, mention their age in discussion, more often share their experiences and motivation to travel, however they do not differ in relation to describing barriers encountered while traveling.
\end{abstract}

\section{Introduction}
According to almost all demographic projections, the coming decades will see a global increase in the population of older adults \cite{ortman2014aging}. What is more, if current trends in societal and technological development hold, this new population of older adults will be healthier, more affluent and mobile than any other older adults cohort in recorded history \cite{sanderson2015faster}, thus they will have more time, resources and abilities to participate in tourism and traveling, both domestic and international, as well as engage in on-line communities. 

According to current research concerning the topic of older adult tourism, and on-line community engagement, these two types of activities provide many positive effects for the older adult population. Tourism can improve overall wellbeing \cite{mitas2012jokes}, as well as the way older adults cope with health issues \cite{ferri2013functional,ferrer2015social}, while community engagement has many positive psychological effects on older adults \cite{sum2009internet}. However, research shows that older adults who wish to participate in life as tourists, experience many barriers and limitations related to their age, and the face other factors both external and internal\cite{kazeminia2015seniors}. Research has shown that addressing these barriers, such as the ones related to safety (as a study by Balcerzak et al. suggest\cite{balcerzak2017press}), can be achieved through listening to the experiences of older adults and applying them in intelligent solutions

With modern developments in fields such as data mining and natural language processing, a new promise opens for addressing the needs of senior travelers directly, through listening, and systematic extracting what they have to say, when spontaneously engaging in activities within on-line travel communities. 

However before designing any forms of automated knowledge extraction, one needs to explore what kind of themes and topics are prevalent in the analyzed corpus of text. By doing so one is able to recognize what specific parts of the texts should be extracted, and what kind of ontology one needs to build, in order to properly represent the structure of the data hidden in a raw text. In this paper we wish to present the results of such an exploratory task.
\subsection{Research contribution \& research question}

We wish to focus on the design of the aforementioned framework, and the information about senior tourists on-line it helps to extract. The following detailed research questions are addressed:
\begin{itemize}
\item{\textbf{What dimensions of social and individual activity are present in on-line communities of older adult tourists.}}
\item{\textbf{In what ways, based on the before mentioned dimensions, do senior tourists differ from other on-line tourists communities?}}
\end{itemize}
These research questions can be translated into the following hypotheses:
\begin{itemize}
\item{\textbf{Hypothesis I:} Tourists aged 50+ mention their age more often than members of other groups}
\item{\textbf{Hypothesis II:} Tourists aged 50+ more often than members of other groups speak about their limitations and barriers}
\item{\textbf{Hypothesis III:} Tourists aged 50+ more often than members of other groups will recall their past experiences of traveling.}
\item{\textbf{Hypothesis IV:} Tourists aged 50+ connect their experiences of limitations and barriers with their age.}
\end{itemize}
The choice of these hypotheses is based on a set of presumption derived from the available literature. Works such as done by \cite{lazar2017going} show that older adults who participate in an on-line community often center their identity around their age. Since this work was based mostly on research of bloggers, we wanted to see if it extends to forum posters.

The rest of this paper is organized as follows. In the Related Work section, we present the review of studies of senior tourism, older adult on-line communities, as algorithms used in related tasks of content extraction. Next, the Analysis tools section, describes in detail the dataset, and methodology used by the researchers. The Results section presents obtained results, while the Conclusions and Further Work section presents the ramifications of these results.

\section{Related work}

\subsection{Older adults as tourists - constraints and limitations}
Studies involving touristic activity among older adults can be split into two main categories. First one, deals with describing travel and touristic patterns among older adults, often compared to other age groups. A study done by Collia \cite{collia20032001}, in which travel patterns of older citizens of the US were analyzed, would be a prime example. Other examples include work by Nyaupane \cite{nyaupane2008seniors}. Such general overviews of travel patterns by older adults were not limited to the US, but were also attempted among other places in Poland by Omelan \cite{omelan2016tourist} and China, by Gu \cite{Gu_2016}. Both studies focused on differences within the older adults demographic, with the study by Omelan finding differences based on place of residence (rural vs Urban), and the study by Gu analyzing through logistic regression the impact of well-being, gender and social status on older adults travel patterns. Studies in this category show also often a positive correlation between various measures of psychological and physical well-being and traveling, like in studies by Ferri \cite{ferri2013functional} and Ferrer \cite{ferrer2015social}.

Another line of research in this category are studies concerning the perception of traveling among older adults. A study done by Gao \cite{gao2016never}, which focuses on travel perception among older Chinese women, is a good example. Research was also done with regard to emotional aspects of traveling, for example a study by Mitas \cite{mitas2012jokes} focused on the role of positive emotion and community building among senior tourists, while a study by Zhang \cite{zhang2016too} described the constraints related to visiting places connected to historical atrocities.

The other category of studies focuses more on researching the constraints faced by older adults who partake in traveling. The prime example of such research, which served as an inspiration for the framework presented in this paper, was work done by Kazeminia  \cite{kazeminia2015seniors} where a dedicated tool for content analysis called THE LEXIMANCER \textit{http://info.leximancer.com/} was used on a corpus of posts made by older adults on the Trip Advisor web page, to extract a framework of constraints faced by older adults who travel. Other exploratory work in this regard was done by Kim \cite{kim2015examination}, where three main types of constraints were identified, the intrapersonal, interpersonal, and structural ones. A similar study done by Nimrod \cite{nimrod2012online} revealed the most persistent constraints through content analysis. However, this study was criticized by authors like Mitas \cite{mitas2012jokes} for failing to focus on the emotional aspects of the older adults traveling community.

The topic of travel constraint among older adults was also approached by Bacs \cite{bacs2016common}, and Fleisher \cite{fleischer2002tourism} as well as, Gudmundsson \cite{gudmundsson2016challenges}, Reed \cite{Reed2007687} and Page \cite{page2015developing} who focused more on health related issues.

\subsection{Older adults and on-line communities}
Substantial work in this topic was done by Nimrod who not only managed to conduct quantitative analysis in order to discover specific clusters within the older adult communities (information swappers, aging-oriented, socializers) \cite{nimrod2013probing}, extracted, through content analysis tools, topics mostly discussed in senior on-line forums \cite{nimrod2010seniors}, but also, through on-line surveys identified psychological benefits of participating in on-line communities \cite{nimrod2014benefits}. There were also studies, like the one by Kowalik and Nielek which focused on communities of older programmers \cite{kowalik2016senior}.

Another thread of research was focused on barriers concerning participation in on-line communities. Research like the one done by Gibson \cite{gibson2010designing} and Jaeger \cite{jaeger2008developing} states that issues of privacy and accessibility form one of the biggest barriers. Noteworthy is also research done by Lazar \cite{lazar2017going} who, by analyzing blog posts of bloggers aged 50+, showed how people within the community of older adults can address their own identity and combat ageist stereotypes.

As participating in on-line communities is crucial for improving well-being of older adults many researchers tried to animate such activities \cite{kopec2017livinglab,kopec2017location,nielek2017turned}.

\subsection{Age, aging, ageism}
Another factor important for the design of a framework for understanding the experiences of senior tourists is the perception of older adults' age, by themselves, the older adults, as well as from the perspective of others. 

In the context of senior tourism, research, such as the one done by Cleaver \cite{cleaver2002want} shows that the perception of ones age is in a mutual relation with ones particular motivations for travelling. Work done by Ylanne \cite{ylanne-mcewen_1999}, on the other hand, represents the role of age perception among travel agencies as an important factor.

\subsection{Automated information extraction in the tourist domain}
With the current availability of on-line tourist reviews, research developed into the application of text mining and data mining solutions that would make use of this data. Authors have already mentioned the recent work by Gu \cite{Gu_2016} in which travel tips would be extracted from travel reviews. Other recent threads of research in this field are represented by Bhatnagar, who also proposed a framework for extracting more subtle, albeit more general and simpler aspects of travel experiences, and a study concerning a methodology of mapping the data related to medical tourism in South Korea done by \cite{huh2017method}. 

\section{Study design}
\subsection{Collected data}
In order to design the framework and develop its applications, we decided to collect textual data from a single source related, which is the 'Older traveler' forum on the Lonely Planet website.

In total, over 30 thousand posts from the last 15 years of the forum’s existence were collected into a single database to serve as a base for future testing and annotating jobs. When compared to previous attempts at content analyzing tourism among older adults such as works by \cite{kazeminia2015seniors}, this dataset provides a definitely larger scope of data to be collected and analyzed due to its sheer size. However, since the paper presented here focuses on preliminary work in preparing an annotation protocol, only a fraction (100 posts) will be used for the current coding and content analysis tasks. This fraction, like all of the forum posts used in this study, were selected randomly with a use of a simple random sampling.

The choice of this particular site and section of a site dedicated only to older adults was based on methodological considerations. Lonely Planet is one of the most popular websites dedicated to tourism, based on a popular series of guide books for tourists. Within the site one can find various forums dedicated to specific locations, as well as specific demographics groups of travelers, such as older adults, women and members of the LGBTQ community.

Selecting the forum was based on established precedent within the research of on-line communities, like the one done by Nimrod \cite{nimrod2010seniors}, and Berdychevsky \cite{berdychevsky2015let}. Using a dedicated older adult forum, instead of all posts produced by older adults within the forum, was a deliberate choice on the part of the authors. Instead of using an arbitrary age for threshold of defining someone as an older adult, used used self-labeling in form of participation in the older adults forums as a defining point for being treated as an older adult. The collected data bears out this assumption, since all of the posts are written on the behalf of the users, we did not observe a situation when, a younger adult would write on a behalf of the older adults. Since the main focus of this paper is to explore the topic ontology for future extracting and knowledge mapping of older adults as a community, it was important to collect posts from a source were such community exists. Moreover, by obtaining data from a single thematic forum, authors gained an opportunity to include such aspects of community life as emotional stances, as well as arguments used in discussion. These two constitute an important element in the design of the future framework, as well as differentiate the research work presented here, from the established corpus of literature where usually barriers, identity and emotions were researched in isolation.

In order to address the research questions presented at the beginning of this article, we decided to collect additional sources from other Lonely Planet forums. In total 17 thousand posts were extracted, out of which 100 were designated for the coding task. This set serves as a baseline for comparing the older adults against the general population of travelers. out of this set a subset of 100 posts was drawn with the use of simple random sampling.

In total 200 posts will be used for the analysis, such dataset fits perfectly within the standard dataset size used in studies related to the topic \cite{lin2004representation}. This number of posts also ensures that the results of further statistical analysis will be valid as statistical significance is concerned.

\subsection{Literature review}
In order to verify the dimensions of the framework, a literature review was conducted in accordance to the guidelines described by \cite{lin2004representation}. First, a v was conducted in order to acquire the most persistent themes in regards to experiences of older adults on-line in general, and senior travelers in general. The thus extracted themes are presented in APPENDIX I.

\begin{table*}[h!]
\centering
\caption{Code book}
\label{Code_book}
\begin{tabular}{| p{1,5cm} | p{5cm} | p{7,5cm} |}
\hline

Dimension & Definition & Examples \\
\hline
Age Reference & All instances of referring to age either of oneself or of somebody else. Age used as an argument in discussion or mentioned as a part of a narrative. Includes both literal and figurative mentions of age. &  \textit{\tabitem{As a 70 year old...  For people my age} \tabitem{In your age it is important} \tabitem{A senior has different needs} \tabitem{I am getting to old for this}}\\
\hline
Location reference & All instances of referring to a specific location, which the user has visited, wants to visits, considers visiting, asks for advice about. Locations include whole continents, countries, cities, provinces, villages or particular locations like tourist attractions, churches, building, land marks etc. & \textit{\tabitem{Has anyone been to Romania?} \tabitem{Visit Singapore, it’s awesome.} \tabitem{I am off to Cambodia} \tabitem{I was In China in the seventies}}\\
\hline
Limitations & This dimension refers to instances of external factors which, in the perception of the poster hinder the individuals ability to travel, or may provide difficulties while traveling to a specific location. Includes both reports of such factors, complaints about such limitations, as well as advice and asking for advice on the topic of external limitations. External factors, are related to traits typical to the environment, rather than the individual.& \textit{\tabitem{ATMs don’t work in Somalia, it’s a pain in the ass}
\tabitem{I couldn’t go to Vietnam, because my visa expired}
\tabitem{I can’t stay abroad, because I can lose my country’s pension}
\tabitem{I didn’t feel good in a country which treats women badly} 
\tabitem{Use a visa card, and exchange your dollars to a local currency, it is better that way}
\tabitem{Ulaanbatar has only one working ATM}
\tabitem{The local police is corrupt, they wouldn't help me unless I bribed them}
\tabitem{Any advice on dealing with altitude sickness in Peru?}}
\\
\hline
Ageism & Refers to any instance using ageist stereotypes in discussion, or a description of ageist experiences held by the user.& \textit{\tabitem{You old people are all the same with your fear} \tabitem{When I was standing in line for the cruise, some young fellow told me that in my age, I am not fit for ocean cruises to Antarctica}}\\
\hline
Barriers & This dimension refers to instances of internal factors which, in the perception of the pster, hinder the individuals ability to travel, or may provide difficulties while traveling to a specific location. Includes both reports of such factors, complaints about such limitations, as well as advice and asking for advice on the topic of internal barriers. These are typical to the individual. & \textit{\tabitem{I would love to travel, but my husband doesn’t allow it}
\tabitem{The fact that I have to work limits my travel plans}
\tabitem{This year I skipped South Asia, because my mother died}
\tabitem{I wish I could go backpack, but with my sick knees it became impossible}
\tabitem{I was afraid of malaria in India}}
\\
\hline
Experiences & Refers to experiences of traveling, recounting of encounters made during ones travels, regardless of location and time perspective. This includes particular experiences, as well as general observations taken from the single experiences.&\textit{ \tabitem{When we entered the market, I saw elephants and arabian spices on display}
\tabitem{The cruise was amazing!}
\tabitem{The hostels I’ve been to always were full of interesting people of all ages}}
\\
\hline
Motivation & Refers to experiences of traveling, recounting of encounters made during ones travels, regardless of location and time perspective. This includes particular experiences, as well as general observations taken from the single experiences. & \textit{\tabitem{I would love to visit India, because I want to spiritually awake}
\tabitem{Cambodia is great if you want escape from the fast paced life of western cities}
\tabitem{This is why I travel, this gives me a sense of being alive}
\tabitem{I really love to travel solo, it lets me fully enjoy my experiences}}
\\
\hline
\end{tabular}
\end{table*}

\subsection{Coding}
The coding job was initiated with the objective of marking the entire corpus. In order to conduct the task, dedicated software called QDA MINER LITE (version 2.0.1). A group of annotators was selected for the task and instructed in the tagging protocol. Each post was read by two annotators who were to mark whether or not the extracted themes appeared in a given post. During the coding task the annotators would discuss any discrepancies in their tagging. The conclusions of this discussion were later applied to the code book. After the coding job was finished and all discrepancies would be addressed, tagging annotation agreement was calculated with the use of Cohen's Kappa, the value achieved (83\%) indicated a high level of inter-rater agreement.

\begin{table*}[h]
\label{tab_freq}
\centering
\caption{Frequencies of codes in the collected samples}
\begin{tabular}{|p{2,5cm}|p{1,2cm}|p{1,2cm}|p{1,2cm}|p{1,15cm}|p{1,2cm}|p{1,2cm}|p{1,2cm}|}
\hline
Group & Age reference & Location reference & Ageism & Limi-
tations & Barriers & Exper
-iences & Moti-
vation \\
\hline
Senior tourist & 20\% & 39\% & 6\% & 22\% & 9\% & 37\% & 15\% \\
\hline
General population & 2\% & 42\% & 0\% & 20\% & 1\% & 20\% & 4\% \\
\hline
\end{tabular}
\newline
Since more than one dimension could occur in a single post, the percentages do not add up to 100\%.
\end{table*}
\section{Results}

Due to the nature of the tagged corpora, we decided to use two statistical measures for testing the validity of the claimed hypotheses. First measure, the T-student's test was used for testing the significance of the observed differences between seniors and general population. Therefore, the hypotheses I to III can be translated to a set of statistical hypotheses about the average difference between posts from two independent samples. For the sake of further testing, the significance level for all of the tests has been set to 0,05. However, due to the fact that multiple pairs of values are compared at once, a correction was applied to the significance threshold, based on the Bonferroni correction. The adjusted significance threshold was therefore set to 0,007. Hypothesis IV, on the other hand, will be addressed with the use of Cohen's kappa as a measure of agreement between two dimensions in a single post.

The frequency of the dimensions present in the tagged posts, is shown in table \ref{tab_freq}. The values given in the table present how often within all of the posts from a specific group did a given dimension occur. Since more than one dimension could occur in a single post, the percentages do not add up to 100\%. Even without conducting the statistical tests, one can notice some general trends visible in the data. First, the references to location and personal experiences are the most frequent aspects of the framework, visible in all three groups. There is also a stark contrast between the references of external and internal hindrances. While limitations, for the most part, are mentioned quite often, the barriers are scant, especially outside of the senior tourist group.

\begin{table*}[h]
\centering
\label{T_test}
\caption{T-test scores}
\begin{tabular}{|p{3cm}|p{1,2cm}|p{1,2cm}|p{1,2cm}|p{1,15cm}|p{1,2cm}|p{1,2cm}|p{1,2cm}|}
\hline
Older adults compared to: & Age reference & Location reference & Ageism & Limi-
tations & Barriers & Exper
-iences & Moti-
vation \\
\hline
General population& 4.24**& -0.43 & 2.52** & -0.34 & 2.64** & 2.7** & 2.7**\\
\hline
\end{tabular}
\newline
*significant difference at a=0,05 \newline
*significant difference at a=0,007
\end{table*}

Intergroup comparison provides further insight into the differences between the designed dimension. By comparing the occurrences of the age reference and ageism between the senior tourists and the general population, one can clearly see that age references within the Lonely Planet forums are almost exclusively related with the senior tourist community, which confirms the validity of Hypothesis I. It has to be pointed out, however, that experiences of ageism are reported rare (only 6\% of the posts). This may mean that explicit forms of ageism are not the main facet of the older adult identity, unlike more subtle implicit forms such as internalized ageism which may be expressed by age references in a specific context. This confirms the findings on the language of ageism made by \cite{gendron2015language}, and provide an additional topic for developing algorithms for text mining.

The relation between age, limitations and barriers expressed in Hypothesis II, when analyzed with the given data also provides interesting results. Although there is no significant difference between senior tourists and general population groups when it comes to mentioning limitations, the difference between mentioning barriers is visible and significant, even tough it is a very scarcely appearing dimension. This lends credence to the notion that older adults who are tourists may experience a varying set of problems when engaging in their travels. However, in view of these results, the hypothesis was validated only partly within the dimension of internal factors such as health and personal issues. This lines with results obtained by other researchers \cite{kazeminia2015seniors}. 

Hypothesis IV stated that older adult tourists, when discussing their limitation and barriers, associate them with their age. In order to validate this hypothesis, authors decided to calculate the Cohen's kappa agreement index between the age reference and limitations and barriers dimensions within the senior tourist group. The obtained result of Cohen's Kappa was 0 in both cases, which indicates that the relation between the given dimension is no different from the one that would arise from pure chance. The ramifications of this observation are crucial. It shows that there is no direct, explicit relation between the perception of age and factors that hinder the older adults plans for travel. Therefore, Hypothesis IV has to be deemed as falsified. However, this finding corroborates the observations presented when analyzing Hypothesis II.

Hypothesis III focused on recalling the traveling experiences of older adults. As shown in table II, it is clear that tourists from the senior tourist group are significantly more likely to recall their travel experiences when engaging within their on-line community. This leads to the conclusion that third hypothesis is validated.
\section{Conclusions \& Future work}
By conducting the literature review and a coding task, supplemented by a the analysis of codes, authors planned to better understand what aspects of the dynamics of senior tourist on-line communities are persistent and demand more focus when designing a framework for the automated mapping and extraction of knowledge from unstructured text created by an on-line community. The analysis of data leads the authors to accept non trivial conclusions in regards to the nature of on-line communities of older adult tourists.

First of all, the role of age or the perception of ones age exists in a complex relation with other aspects of the community. Even though the perception of age is an important part of the community of older adult tourists, it is not, at least when explicitly stated, related to the problems they face while traveling. What is more, there is no difference between mentions of external limitations between older adults and general users of the travel forum, which means that older adults face problems at similar rates to other groups.

What sets them aside, however, is their tendency to report problems related to health and personal issues. Moreover, older adults are more likely to share their experiences to the travels they made.


These findings have significant ramifications for the knowledge extraction projects as proposed in the Introduction to this article. The fact that older adults do not verbally associate their problems with their age, indicated that the machine learning algorithms would have to be focused not only on detecting the dimensions of limitations experienced by older adults, but also should be able to extract implicit relations between them and the age of tourists.

The importance of storytelling and motivation for older adults also provides an additional level of challenge for the framework. Here the use of such algorithms like recursive neural networks is suggested by the authors to properly extract the sequence of events described by the poster, while the motivation aspect can be connected to the limitations and age dimensions, so that a more comprehensive picture of older adults experiences can be extracted.

What is more, the task of extracting knowledge not explicitly stated by the members of the on-line community would constitute an interesting topic in the field of natural language processing. 

The full corpus of texts from the traveling forums will be tagged by a group of trained taggers and later will be used for training a set of machine learning algorithms in order to automatically detect the designed dimensions of the framework, as well as mapping the relations between them. The database thus created would connect specific locations with the specific problems and experiences typical to the members of the 50+ demographic category. In the final stage the framework would be embedded in a web application that would allow us to extract structured implicit knowledge that can be learned from the communities of older adult tourists.
Such algorithms can also be combined with knowledge on developer team work in order to further improve its quality\cite{turek2011wikiteams},\cite{wierzbicki2010learning}.
\section{Acknowledgments.}

This project has received funding from the European Union’s Horizon 2020 research and innovation programme under the Marie Skłodowska-Curie grant agreement No 690962.

\bibliographystyle{splncs03.bst}
\bibliography{new_bib}

\begin{thebibliography}{10}
\providecommand{\url}[1]{\texttt{#1}}
\providecommand{\urlprefix}{URL }

\bibitem{balcerzak2017press}
Balcerzak, B., Kope{\'c}, W., Nielek, R., Kruk, S., Warpechowski, K., Wasik,
  M., W{\k{e}}grzyn, M.: Press f1 for help: participatory design for dealing
  with on-line and real life security of older adults. arXiv preprint
  arXiv:1706.10223  (2017)

\bibitem{bacs2016common}
Ba{\c{s}}, M.: Common constraints of tourism market segments: Examples of
  senior tourism and disabled tourism. In: International Conference on Tourism
  Dynamics and Trends, proceedings book. pp. 54--73 (2016)

\bibitem{berdychevsky2015let}
Berdychevsky, L., Nimrod, G.: " let's talk about sex": Discussions in seniors'
  online communities. Journal of Leisure Research  47(4),  467 (2015)

\bibitem{cleaver2002want}
Cleaver, M., Muller, T.E.: I want to pretend i’m eleven years younger:
  Subjective age and seniors’ motives for vacation travel. In: Advances in
  Quality of Life Research 2001, pp. 227--241. Springer (2002)

\bibitem{collia20032001}
Collia, D.V., Sharp, J., Giesbrecht, L.: The 2001 national household travel
  survey: A look into the travel patterns of older americans. Journal of safety
  research  34(4),  461--470 (2003)

\bibitem{ferrer2015social}
Ferrer, J.G., Sanz, M.F., Ferrandis, E.D., McCabe, S., Garc{\'\i}a, J.S.:
  Social tourism and healthy ageing. International Journal of Tourism Research
  (2015)

\bibitem{ferri2013functional}
Ferri, M., Dur{\'a}, E., Garc{\'e}s, J.: Functional health benefits for elderly
  people related to social tourism policy promotion. International Journal of
  Multidisciplinary Social Sciences  1,  1--8 (2013)

\bibitem{fleischer2002tourism}
Fleischer, A., Pizam, A.: Tourism constraints among israeli seniors. Annals of
  Tourism Research  29(1),  106--123 (2002)

\bibitem{gao2016never}
Gao, J., Kerstetter, D.: Never too old to travel: Exploring older chinese
  women’s perceptions of travel. Paper presented at the Tourism Travel and
  Research Association 2016 Conference (2016)

\bibitem{gendron2015language}
Gendron, T.L., Welleford, E.A., Inker, J., White, J.T.: The language of ageism:
  Why we need to use words carefully. The Gerontologist p. gnv066 (2015)

\bibitem{gibson2010designing}
Gibson, L., Moncur, W., Forbes, P., Arnott, J., Martin, C., Bhachu, A.S.:
  Designing social networking sites for older adults. In: Proceedings of the
  24th BCS Interaction Specialist Group Conference. pp. 186--194. British
  Computer Society (2010)

\bibitem{Gu_2016}
Gu, D., Zhu, H., Brown, T., Hoenig, H., Zeng, Y.: Tourism experiences and
  self-rated health among older adults in china. Journal of Aging and Health
  28(4),  675--703 (2016), pMID: 26486781

\bibitem{gudmundsson2016challenges}
Gudmundsson, A., Stevenson, J., Petrovic, M., Somers, A., Onder, G., Callens,
  S., van~der Cammen, T.J.: Challenges and risks for older travellers with
  multimorbidity: Focus on pharmacotherapy. European Geriatric Medicine  7(5),
  407--410 (2016)

\bibitem{huh2017method}
Huh, J.H., Kim, H.B., Kim, J.: A method of modeling of basic big data analysis
  for korean medical tourism: A machine learning approach using apriori
  algorithm. In: International Conference on Information Science and
  Applications. pp. 784--790. Springer (2017)

\bibitem{jaeger2008developing}
Jaeger, P.T., Xie, B.: Developing online community accessibility guidelines for
  persons with disabilities and older adults. Journal of Disability Policy
  Studies pp. 55--63 (2008)

\bibitem{kazeminia2015seniors}
Kazeminia, A., Del~Chiappa, G., Jafari, J.: Seniors’ travel constraints and
  their coping strategies. Journal of Travel Research  54(1),  80--93 (2015)

\bibitem{kim2015examination}
Kim, H.L.: An Examination of Salient Dimensions of Senior Tourist Behavior:
  Relationships among Personal Values, Travel Constraints, Travel Motivation,
  and Quality of Life (QoL). Ph.D. thesis (2015)

\bibitem{kopec2017location}
Kope{\'c}, W., Abramczuk, K., Balcerzak, B., Ju{\'z}win, M., Gniadzik, K.,
  Kowalik, G., Nielek, R.: A location-based game for two generations: Teaching
  mobile technology to the elderly with the support of young volunteers. In:
  eHealth 360°, pp. 84--91. Springer International Publishing (2017)

\bibitem{kopec2017livinglab}
Kope{\'c}, W., Skorupska, K., Jaskulska, A., Abramczuk, K., Nielek, R.,
  Wierzbicki, A.: Livinglab pjait: Towards better urban participation of
  seniors. arXiv preprint arXiv:1707.00030  (2017)

\bibitem{kowalik2016senior}
Kowalik, G., Nielek, R.: Senior programmers: Characteristics of elderly users
  from stack overflow. In: International Conference on Social Informatics. pp.
  87--96. Springer International Publishing (2016)

\bibitem{lazar2017going}
Lazar, A., Diaz, M., Brewer, R., Kim, C., Piper, A.M.: Going gray, failure to
  hire, and the ick factor: Analyzing how older bloggers talk about ageism. In:
  Proceedings of the 2017 ACM Conference on Computer Supported Cooperative Work
  and Social Computing. pp. 655--668. ACM (2017)

\bibitem{lin2004representation}
Lin, M.C., Hummert, M.L., Harwood, J.: Representation of age identities in
  on-line discourse. Journal of Aging Studies  18(3),  261--274 (2004)

\bibitem{mitas2012jokes}
Mitas, O., Yarnal, C., Chick, G.: Jokes build community: Mature tourists’
  positive emotions. Annals of Tourism Research  39(4),  1884--1905 (2012)

\bibitem{nielek2017turned}
Nielek, R., Lutostanska, M., Kopec, W., Wierzbicki, A.: Turned 70? it is time
  to start editing wikipedia. arXiv preprint arXiv:1706.10060  (2017)

\bibitem{nimrod2010seniors}
Nimrod, G.: Seniors’ online communities: A quantitative content analysis. The
  Gerontologist  50(3),  382--392 (2010)

\bibitem{nimrod2012online}
Nimrod, G.: Online communities as a resource in older adults’ tourism. The
  Journal of Community Informatics  8(1) (2012)

\bibitem{nimrod2013probing}
Nimrod, G.: Probing the audience of seniors’ online communities. The Journals
  of Gerontology Series B: Psychological Sciences and Social Sciences  68(5),
  773--782 (2013)

\bibitem{nimrod2014benefits}
Nimrod, G.: The benefits of and constraints to participation in seniors’
  online communities. Leisure Studies  33(3),  247--266 (2014)

\bibitem{nyaupane2008seniors}
Nyaupane, G.P., McCabe, J.T., Andereck, K.L.: Seniors' travel constraints:
  Stepwise logistic regression analysis. Tourism Analysis  13(4),  341--354
  (2008)

\bibitem{omelan2016tourist}
Omelan, A., Podstawski, R., Raczkowski, M.: Tourist activity of senior citizens
  (60+) residing in urban and rural areas. Physical Culture and Sport. Studies
  and Research  72(1),  51--65 (2016)

\bibitem{ortman2014aging}
Ortman, J.M., Velkoff, V.A., Hogan, H., et~al.: An aging nation: the older
  population in the united states. Washington, DC: US Census Bureau pp.
  25--1140 (2014)

\bibitem{page2015developing}
Page, S.J., Innes, A., Cutler, C.: Developing dementia-friendly tourism
  destinations: an exploratory analysis. Journal of Travel Research  54(4),
  467--481 (2015)

\bibitem{Reed2007687}
Reed, C.M.: Travel recommendations for older adults. Clinics in Geriatric
  Medicine  23(3),  687 -- 713 (2007),
  \url{http://www.sciencedirect.com/science/article/pii/S0749069007000407},
  infectious Diseases in Older Adults

\bibitem{sanderson2015faster}
Sanderson, W.C., Scherbov, S.: Faster increases in human life expectancy could
  lead to slower population aging. PloS one  10(4),  e0121922 (2015)

\bibitem{sum2009internet}
Sum, S., Mathews, R.M., Pourghasem, M., Hughes, I.: Internet use as a predictor
  of sense of community in older people. CyberPsychology \& Behavior  12(2),
  235--239 (2009)

\bibitem{turek2011wikiteams}
Turek, P., Wierzbicki, A., Nielek, R., Datta, A.: Wikiteams: How do they
  achieve success? IEEE Potentials  30(5),  15--20 (2011)

\bibitem{wierzbicki2010learning}
Wierzbicki, A., Turek, P., Nielek, R.: Learning about team collaboration from
  wikipedia edit history. In: Proceedings of the 6th International Symposium on
  Wikis and Open Collaboration. p.~27. ACM (2010)

\bibitem{ylanne-mcewen_1999}
YLANNE-McEWEN, V.: ‘young at heart’: discourses of age identity in travel
  agency interaction. Ageing and Society  19(4),  417–440 (1999)

\bibitem{zhang2016too}
Zhang, H., Yang, Y., Zheng, C., Zhang, J.: Too dark to revisit? the role of
  past experiences and intrapersonal constraints. Tourism Management  54,
  452--464 (2016)

\end{thebibliography}

\end{document}